\documentclass[conference]{IEEEtran}
\IEEEoverridecommandlockouts
\usepackage{cite}
\usepackage{amsmath,amssymb,amsfonts}
\usepackage{algorithmic}
\usepackage{graphicx}
\usepackage{textcomp}
\usepackage{xcolor}
\usepackage{subfigure}
\usepackage{multirow} 
\usepackage{makecell}
\usepackage{booktabs}
\usepackage{acronym}
\usepackage{subcaption}
\usepackage{url} 
\usepackage{xcolor}
\usepackage{pifont}
\usepackage{colortbl}
\usepackage{hyperref}

\def\BibTeX{{\rm B\kern-.05em{\sc i\kern-.025em b}\kern-.08em
    T\kern-.1667em\lower.7ex\hbox{E}\kern-.125emX}}
\begin{document}

\acrodef{GANs}[GANs]{Generative Adversarial Networks}
\acrodef{RL}[RL]{Reinforcement Learning}
\acrodef{CLIP}[CLIP]{Contrastive Language-Image Pre-Training}
\acrodef{OKS}[OKS]{Object Keypoint Similarity}
\acrodef{FID}[FID]{Fréchet Inception Distance}
\acrodef{IS}[IS]{Inception Score}
\acrodef{CR}[CR]{Compression Ratio}
\acrodef{IQA}[IQA]{Image Quality Assessment}
\acrodef{Anime}[Anime]{Anime-Style Line Drawing}
\acrodef{Opensketch}[Opensketch]{Opensketch-Style Line Drawing}
\acrodef{GUI}[GUI]{Graphical user interface}

\title{SketchRef: a Multi-Task Evaluation Benchmark for Sketch Synthesis}

\author{
    Xingyue Lin, Xingjian Hu, Shuai Peng, Jianhua Zhu,
    Liangcai Gao\textsuperscript{*}\thanks{*Corresponding author} 
    \\
    Wangxuan Institute of Computer Technology, Peking University, Beijing, China \\
    linxingyue23@stu.pku.edu.cn, \{huxingjian, pengshuaipku, zhujianhuapku, gaoliangcai\}@pku.edu.cn
}

\maketitle

\begin{abstract}
Sketching is a powerful artistic technique for capturing essential visual information about real-world objects and has increasingly attracted attention in image synthesis research. However, the field lacks a unified benchmark to evaluate the performance of various synthesis methods.
To address this, we propose SketchRef, the first comprehensive multi-task evaluation benchmark for sketch synthesis. SketchRef fully leverages the shared characteristics between sketches and reference photos. It introduces two primary tasks: category prediction and structural consistency estimation, the latter being largely overlooked in previous studies. These tasks are further divided into five sub-tasks across four domains: animals, common things, human body, and faces.
Recognizing the inherent trade-off between recognizability and simplicity in sketches, we are the first to quantify this balance by introducing a recognizability calculation method constrained by simplicity, mRS, ensuring fair and meaningful evaluations. To validate our approach, we collected 7,920 responses from art enthusiasts, confirming the effectiveness of our proposed evaluation metrics.
Additionally, we evaluate the performance of existing sketch synthesis methods on our benchmark, highlighting their strengths and weaknesses. We hope this study establishes a standardized benchmark and offers valuable insights for advancing sketch synthesis algorithms.
\end{abstract}


\section{Introduction}
\label{sec:intro}

Sketching, a form of drawing commonly used by human artists, aims to capture the key features of real-world objects through line and contour. 
Despite the sketch being simple, consisting of only a few lines, humans can still recognize the structure and category of the main object from the sketch.

Researchers have recently ventured into deep learning to automate sketch synthesis. 
Compared to human sketching, the automatic synthesis methods~\cite{li2019photo, yi2020unpaired, vinker2022clipasso, muhammad2018learning, radford2021learning, xing2024diffsketcher} can effectively save time and financial costs.
However, there is currently no standardized benchmark for sketch synthesis evaluation, which hinders the further development of this field.
A unified multi-domain dataset has yet to be established, and the evaluation metrics are primarily limited to classification accuracy for measuring the recognizability of sketches.
Additionally, sketching inherently involves simplification, but the trade-off between recognizability and simplicity is seldom considered during evaluation, which makes the comparison of sketch synthesis algorithms unfair.

\begin{table}[t]    
   \caption{Comparison of evaluation tasks and metrics used in different synthesis methods: Clipasso~\cite{vinker2022clipasso}, Clipascene~\cite{vinker2023clipascene},  LineDrawings~\cite{chan2022learning} and UPDG~\cite{yi2020unpaired}. Previous methods overlook the task and metrics of structural consistency, fail to consider the trade-off between recognizability and simplicity, and limit the calculation of simplicity to the number of strokes.}
\label{tab:metric overview}
\centering
\begin{tabular}{cccccc}
\toprule
\multirow{2}{*}{\textbf{Metric}} & \multirow{2}{*}{\textbf{Aspects}} & \multicolumn{4}{c}{\textbf{Methods}}      \\ \cline{3-6} &  & \cite{vinker2022clipasso}, \cite{vinker2023clipascene} & \cite{chan2022learning} & \cite{yi2020unpaired} & \textbf{OURS} \\ \midrule
\multirow{2}{*}{Recognizability} & category                          & {\ding{51}}                                           & {\ding{51}}            & {\ding{51}}         & {\ding{51}}  \\
                                 & structure                         & {\ding{55}}                                                       & {\ding{55}}                        & {\ding{55}}                      & {\ding{51}}  \\ \midrule
\multirow{2}{*}{Simplicity}      & stroke                      & {\ding{51}}                                           & {\ding{55}}                        & {\ding{51}}          & {\ding{51}}  \\
                                 & pixel                       & {\ding{55}}                                                       & {\ding{55}}                        & {\ding{55}}                      & {\ding{51}}  \\ \midrule
Overall                          & trade-off                          & {\ding{55}}                                                       & {\ding{55}}                        & {\ding{55}}                      & {\ding{51}}  \\ \bottomrule
\vspace{-5mm}
\end{tabular}
\end{table}

To fill this gap, we introduce \textbf{SketchRef}, a multi-task benchmark for sketch synthesis evaluation, based on the shared features between the \textbf{Sketch} and the \textbf{Ref}erence photo.
We collect data from diverse domains, including humans, faces, animals, and common things. Given their varying characteristics, we design two primary tasks: category prediction for animal and thing data, and structural consistency estimation for human, face, and animal data. The latter leverages shared structural features, such as key point alignment (e.g., eyes and limb joints), between sketches and reference photos.
To the best of our knowledge, we are the first to propose the task of estimating shared structure, as shown in Table~\ref{tab:metric overview}.
\begin{figure*}[htbp]
  \centering
  \begin{minipage}{0.48\textwidth}
    \centering
    \label{fig:overview:structure} 
    \includegraphics[width=\textwidth]{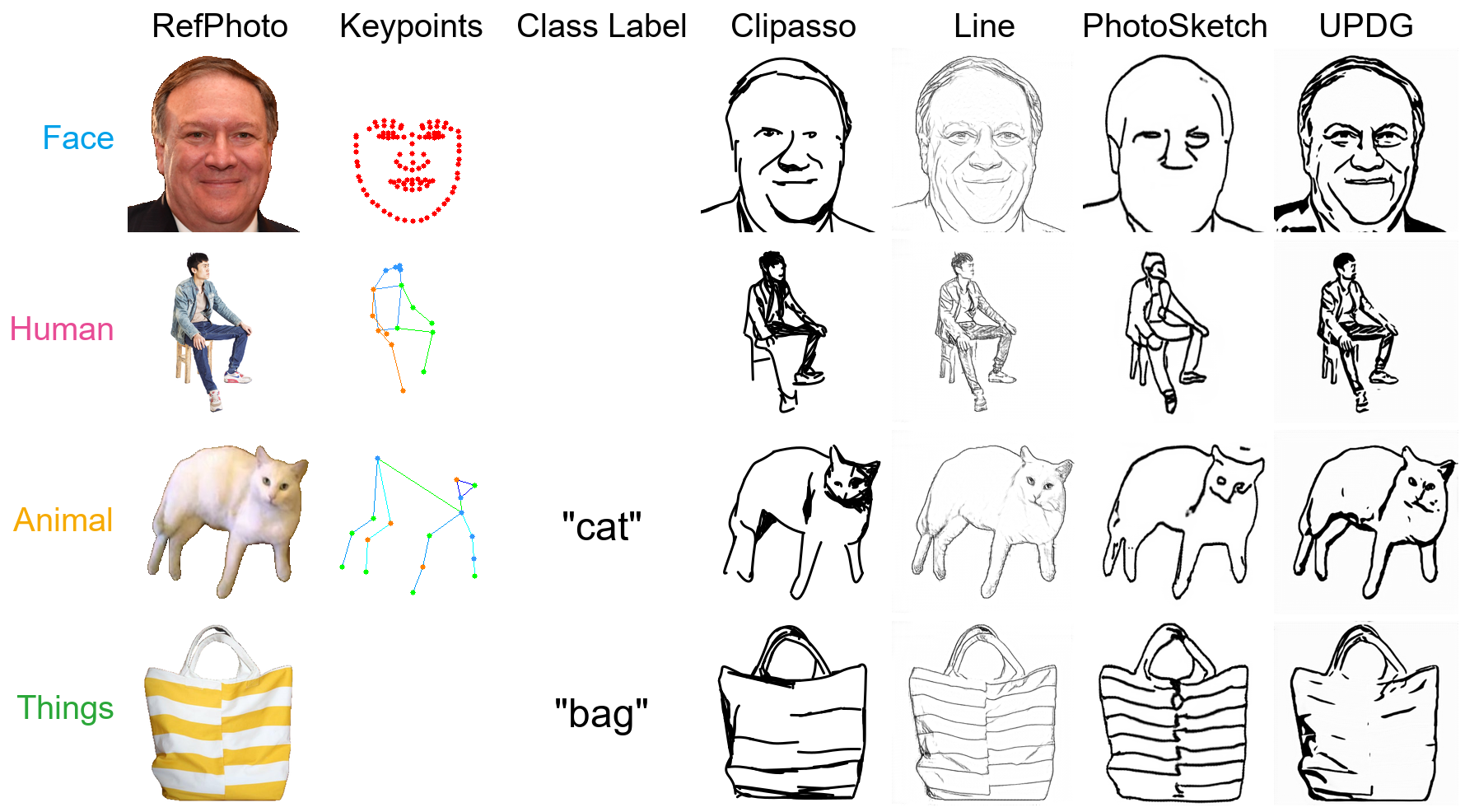}
  \end{minipage}
  \hfill
  \begin{minipage}{0.48\textwidth}
    \centering
    \resizebox{0.95\textwidth}{!}{%
    \begin{tabular}{cccccccc}
    \toprule
    \multirow{2}{*}{\textbf{Task}} & \multirow{2}{*}{\textbf{Domain}} & \multicolumn{6}{c}{\textbf{Methods}}                                                                                                        \\ \cline{3-8} 
                                   &                                  & \cite{vinker2023clipascene} & \cite{li2019photo} & \cite{chan2022learning} & \cite{muhammad2018learning} & \cite{vinker2022clipasso} & \textbf{OURS} \\ \midrule
    \multirow{2}{*}{Category}      & Animal                           & {\ding{51}}                                   & {\ding{51}}       & {\ding{51}}            & {\ding{51}}      & {\ding{51}}          & {\ding{51}}  \\
                                   & Things                           & {\ding{51}}                                   & {\ding{51}}       & {\ding{51}}            & {\ding{51}}    & {\ding{51}}            & {\ding{51}}  \\ \midrule
    \multirow{3}{*}{Structure}     & Animal                           & {\ding{55}}                                               & {\ding{55}}                   & {\ding{55}}                        & {\ding{55}}         & {\ding{55}}                   & {\ding{51}}  \\
                                   & Face                             & {\ding{55}}                                               & {\ding{55}}                   & {\ding{55}}          & {\ding{55}}              & {\ding{55}}                            & {\ding{51}}  \\
                                   & Human                            & {\ding{55}}       & {\ding{55}}                                        & {\ding{55}}                   & {\ding{55}}                        & {\ding{55}}                            & {\ding{51}}  \\ \midrule
   \multicolumn{2}{c}{Size of Dataset}      &35 & 100  & 100 & 200 & 200 &\textbf{4537}\\ \bottomrule
    \end{tabular}
    }
  \end{minipage}
  \caption{Overview of our proposed dataset. The left image shows the data and annotations we cover, as well as sketches synthesized from our data. It can be observed that some of the synthesized sketches miss important structures. For example, in the face sketch synthesized by PhotoSketch~\cite{li2019photo}, the eyebrows and mouth are missing. In the human sketch synthesized by CLIPasso~\cite{vinker2022clipasso}, the right leg is absent. We use keypoints as a bridge to quantify these structural errors. The right table compares the evaluation datasets used in our benchmark method with those of previous methods. It can be seen that our dataset covers a wider range of domains and includes a significantly larger volume of data.}
  \label{fig:overview} 
\vspace{-3mm}
\end{figure*}

In particular, we observe previous image similarity metrics~\cite{wang2004image, zhang2018unreasonable} are unsuitable for the task of structural consistency estimation, unable to capture missing or erroneous key structural information.
To address this issue, we design a new metric based on pose estimation to evaluate the visual structural consistency between sketches and reference images.
Additionally, we explore the trade-off between recognizability and simplicity in sketches and introduce a new evaluation principle, the mean recognizability under simplification (mRS), for fair comparison.
Furthermore, we collect responses from 198 art enthusiasts regarding the recognizability and simplicity of synthesized sketches. 
Finally, based on our proposed evaluation benchmark, we conduct extensive evaluations of 8 representative sketch synthesis methods. 
The results reveal that most synthesis methods lack robustness at higher simplification levels and that effective category prediction does not ensure structural consistency, emphasizing the need for structure-preserving optimization in future methods.

To summarize, our contributions are as follows:
1) We introduce SketchRef, the first multi-task evaluation benchmark for sketch synthesis, offering datasets and metrics across multiple domains.
2) We are the first to introduce the task of structural consistency estimation and propose a quantitative metric aligned with human perception.
3) We propose a new evaluation principle, mRS, ensuring fairness in evaluating sketches with different levels of simplification.
4) We evaluate 8 widely-used synthesis methods and uncover valuable findings that offer meaningful insights for the sketch research.

\section{Related Work}

\subsection{Evaluation Dataset in Sketch Synthesis.}
There is currently no standardized dataset for sketch synthesis evaluation.
Different study chooses different datasets for assessment.
Clipasso~\cite{vinker2022clipasso} employs 200 randomly selected images from 10 categories within the SketchyCOCO dataset~\cite{gao2020sketchycoco}.
Photosketch~\cite{li2019photo} collects 100 outdoor images sourced from Adobe Stock for evaluation. 
LineDrawing~\cite{chan2022learning} utilizes the MIT-Adobe FiveK dataset~\cite{bychkovsky2011learning} and test data from UPDG~\cite{yi2020unpaired}.
These evaluation datasets largely focus on common things~\cite{li2018universal, yu2016sketch, mukherjee2024seva}, and most of them only annotate sketches with category-level labels.
Besides, there is no differentiation between sketches synthesized from different photo categories, overlooking the distinctive structural features of specific sketch types, such as pose information in human and animal sketches.

\subsection{Evaluation Metrics in Sketch Synthesis.}
Previous works~\cite{vinker2022clipasso, muhammad2018learning, vinker2023clipascene} evaluate sketch recognizability using pre-trained classifiers' classification accuracy. 
However, this approach only assesses sketch quality at the category level, neglecting the visual structural consistency between the sketch and the reference photo.
Additionally, the SEVA study~\cite{mukherjee2024seva}, which collects human sketches drawn over varying time intervals, finds that sketches drawn with more time tend to be visually more complex and are perceived as more recognizable by humans. Although this influence of simplicity on recognizability is acknowledged, previous work overlooks simplicity when evaluating the recognizability of sketches. 


\section{SketchRef}

In this section, we will introduce the benchmark in three parts: the primary tasks (Section~\ref{subsec: task}), the proposed dataset (Section~\ref{subsec: dataset}), and the corresponding evaluation metrics (Section~\ref{sec: metrics}).

\subsection{Task Construction}
\label{subsec: task}
When constructing tasks for sketch evaluation, we must consider which aspects of the sketch need to be assessed. 
Unlike general image generation, sketches use lines to abstract the input reference photo, with color inconsistencies and potentially missing non-essential details. 
Nevertheless, humans can still recognize categories and key structures from the sketch.
We believe that this recognizability is a crucial feature of sketches, which led us to design two primary tasks:
\begin{itemize}
\item \textbf{Category Prediction}: Given the category label of the reference photo, we calculate the text similarity between the synthesized sketch and the label, evaluating the category-level recognizability.

\item \textbf{Structural Consistency Estimation}: Given the reference photo, we estimate the consistency between the synthesized sketch and the reference photo in terms of key structures, evaluating the structure-level recognizability.
\end{itemize}

\subsection{Dataset Construction}
\label{subsec: dataset}
As shown in Fig.~\ref{fig:overview}, our proposed evaluation dataset consists of reference photos required for sketching, and annotations shared between sketches and reference photos, which include both category and structural annotations. 
There are 4 domains: Human, Face, Animal, and Things. 
We argue that keypoints in data, such as those for humans and animals, are generalizable, with well-established annotation guidelines. 
In contrast, datasets involving common objects lack universal keypoint standards but offer rich category information.
Therefore, based on the 4 domains, we define 5 tasks: category prediction for Animal and Things, and structural consistency estimation for Animal, Face, and Human.

For Human, to ensure that the human poses are clearly visible and occupy the central position of the image, we collect 1,137 photos of human models from a free public human figure reference website for artists, including various poses such as standing and squatting. 
We annotate human keypoints using the COCO format~\cite{lin2014microsoft}, identifying 17 points at major joints. 
For Face, we collect 950 face photos from the FFHQ dataset~\cite{karras2019style}, which includes different ages, genders, and ethnicities, with 106 dense keypoints for each target. 
For Animal, we collect 950 photos from the Animal-Pose evaluation dataset~\cite{cao2019cross}, with five animal class labels (dog, cat, cow, horse, and sheep), and 20 keypoints for each target. 
For Things, we collect 1,500 photos from SEVA~\cite{mukherjee2024seva}, including 127 types of class labels, such as bag, car, etc. We segment the reference photos with U2Net~\cite{qin2020u2} to ensure a blank background, allowing synthesis methods to focus on main objects and avoid background interference during evaluation.

\subsection{Evaluation Metrics}
\label{sec: metrics}
\textbf{Category-level Recognizability} refers to the ability of a sketch to be accurately identified as the category of the reference photo.
Following previous works~\cite{vinker2022clipasso, mukherjee2024seva}, we compute the average cosine similarity between the CLIP embeddings of the class names and the sketches~\cite{radford2021learning}.
The category-level recognizability for a single sketch $x^{\text{skt}}$ is calculated as \( \text{R}_{c} \) :
\begin{equation}
\text{R}_{c}(x^{\text{skt}}) = \cos \left( \text{E}_{\text{text}}(class), \text{E}_{\text{image}}(x^{\text{skt}}) \right)
    \label{eq:clip},
\end{equation}
where \(\cos(\cdot)\) is the cosine similarity, \( class \) represents the class name for the sketch, \( \text{E}_{\text{text}(\cdot)} \) is the CLIP text embedding of the class name, \( \text{E}_{\text{image}(\cdot)} \) is the CLIP image embedding of the sketch.


\textbf{Structure-level Recognizability} aims to measure how well a sketch preserves the key structural features of the reference photo. 
We observe that sketches often simplify details, and certain important structural elements may be drawn incorrectly or omitted, as shown in Fig.~\ref{fig:overview}. 
Previous image similarity metrics, such as SSIM and LPIPS~\cite{wang2004image, zhang2018unreasonable} assess structural similarity based on pixel-level or global feature-level comparisons, which are unable to effectively capture the local omissions or errors in the critical structure of sketches. 
To address this limitation, we leverage the characteristic of shared keypoints between reference photos and sketches, and design a new structural similarity metric.
If the keypoints in the sketch closely match those in the reference photo, the structure-level recognizability of the sketch is considered high. We use an open-source top-down pose estimation model to assess the correspondence of keypoints between the sketch and the reference photo. 
Specifically, since the objects in the reference photo and the sketch correspond one-to-one, we first perform object detection on the reference photo, and then predict the keypoint for each detected region in both the sketch and the reference photo. 
We use the average \ac{OKS} of all targets as our structure-level recognizability \( \text{R}_{s}\):
\begin{equation} \text{R}_{s}(x^{\text{skt}}) = \frac1N \sum_{i=1}^N \text{OKS}(y_i, \hat y_i), \label{eq:OKS} \end{equation}
where $N$ is the number of detected targets in the reference photo, and $y_i$ and $\hat y_i$ represent the keypoint predictions for the $i$-th detected object in the reference photo $x^{\text{ref}}$ and the sketch $x^{\text{skt}}$, respectively. 
Experiments in Section~\ref{sec:experiment} show that our evaluation method is more sensitive to the missing key structural features and aligns more closely with human perception. 

\begin{figure}[htbp]
  \centering
  \subfigure[]{
    \label{fig:stroke:vis} 
    \includegraphics[width=0.185\textwidth]{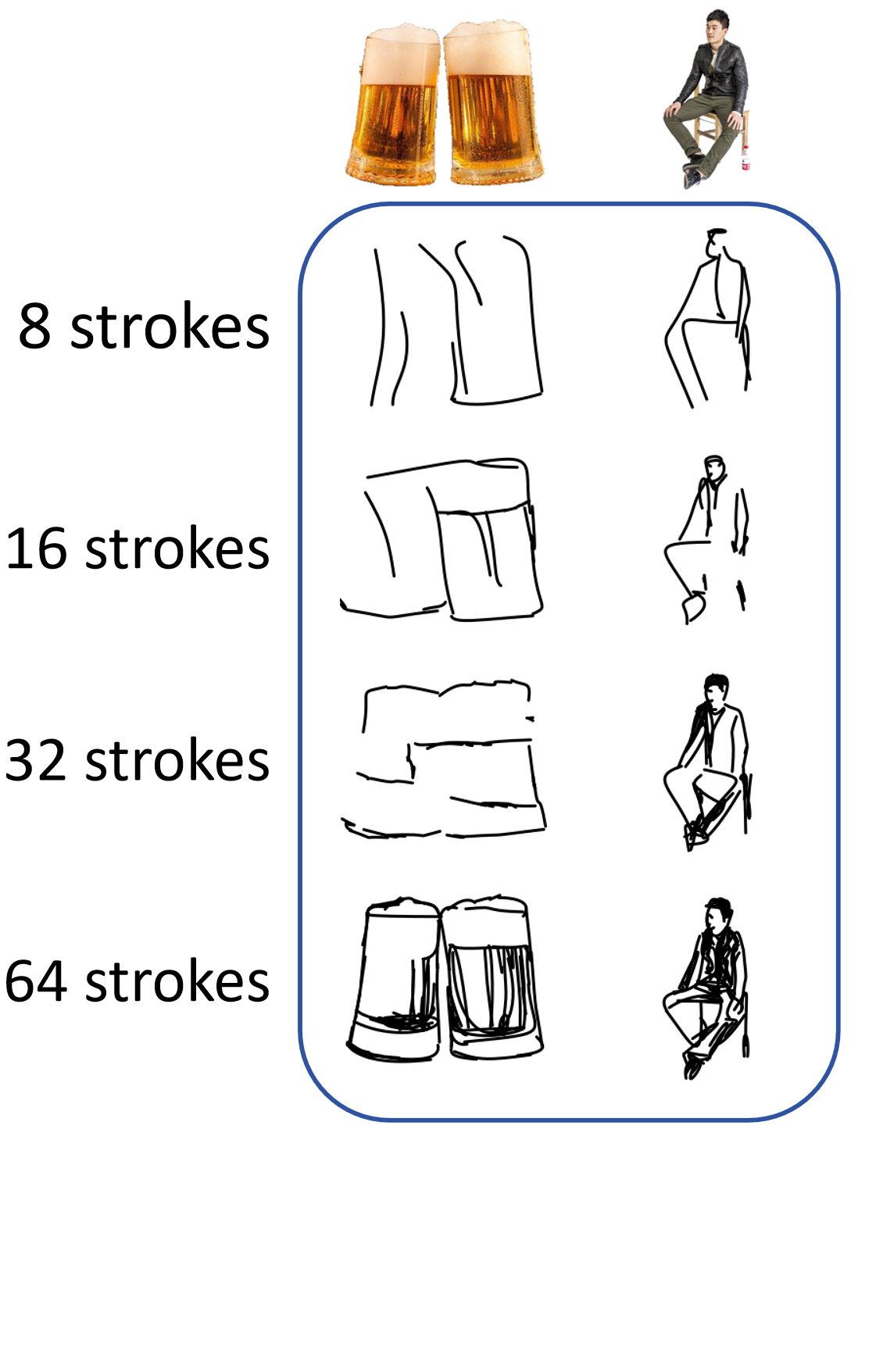}}
  \subfigure[]{
    \label{fig:stroke:column} 
    \includegraphics[width=0.23\textwidth]{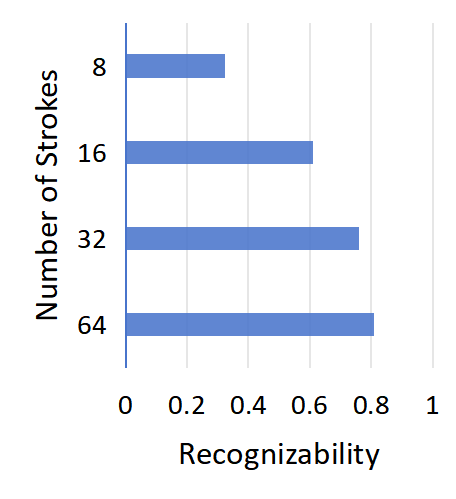}}
  \hspace{0in} 
  \caption{The trade-off between recognizability and simplicity in sketches. (a) Cases of synthesized sketches with different numbers of strokes. (b) Evaluating the value of recognizability using proposed metrics on sketches with varying stroke counts.}
  \label{fig:stroke impact} 
\end{figure}

\textbf{Evaluation Principle.}
Sketching involves simplifying or adding details based on reference photos.
This process inherently involves a trade-off between simplicity and recognizability. 
To validate this trade-off, we use Clipasso~\cite{vinker2022clipasso}, a sketch synthesis method that allows for control over the number of strokes, to synthesize sketches with 8, 16, 32, and 64 strokes. 
As shown in Fig.~\ref{fig:stroke impact}, when the number of strokes increases, the simplicity of the sketch decreases, and recognizability increases. 
If recognizability is calculated without accounting for simplicity, it can lead to unfair comparisons, especially for highly simplified sketches. 
Therefore, it is essential to develop a principle for evaluating recognizability that considers the level of simplification.

One challenge is how to quantify the level of simplification when it comes to pixel images. 
Strokes provide a natural measure of complexity, but this is not the case for pixel images. 
We propose a method to measure simplicity, which involves comparing the sketch $x^{\text{skt}}$ to its reference photo $x^{\text{ref}}$, defined as relative Simplicity Ratio (\( \text{SR} \)):
\begin{equation}
\label{eq:simplicity}
\text{SR}(x^{\text{skt}}) = \frac{\text{C}(x^{\text{ref}})}{\text{C}(x^{\text{skt}})},
\end{equation}
where \(\text{C}(\cdot)\) is a complexity assessment method. 
The value of \( \text{SR} \) ranges from 0 to positive infinity, where \( \text{SR} > 1 \) indicates that the sketch is simpler than the original image, otherwise the opposite.

Building on this simplicity measurement, we propose a new evaluation principle, the mean recognizability under simplification (mRS), which quantifies recognizability constrained by controlled levels of simplification.
For a given set of sketches, we establish a threshold \(\alpha\) for the simplification level. 
We then compute the mRS constrained by the simplification threshold \(\alpha\):
\begin{equation}
\label{eq:mRS}
\text{mRS}@\alpha = \frac{1}{N} \sum_{i=1}^{N} \text{R}(x^{\text{skt}}_i) {\mathbb{I}}(\text{SR}(x^{\text{skt}}_i) > \alpha),
\end{equation}
where \(\text{R}(\cdot)\) is the recognizability value (\(\text{R}_{\text{c}}\) for the category prediction task, \(\text{R}_{\text{s}}\) for the structural consistency estimation task), \( N \) is the total number of sketches, \( x^{\text{skt}}_i \) represents the \(i\)-th sketch, \(\mathbb{I}(\cdot)\) is indicator function. 
We select \(\alpha\) = 0 and \(\alpha\) = 1.5 as two distinct simplification thresholds.
When \(\alpha\) is set to 0, no sketches are filtered out. 
When \(\alpha\) is set to 1.5, sketches with \( \text{SR} \) below 1.5 are filtered out, ensuring that sketches with a sufficient level of simplification remain for further analysis. 
By evaluating each sketch synthesis method's recognizability at controlled simplification levels, we ensure a fair comparison across methods.

\section{Experiment}
\label{sec:experiment}

\subsection{Experimental Setups}

\textbf{Datasets.}
To evaluate sketch synthesis methods, we use our proposed evaluation dataset.
All images used for synthesis and evaluation are 224x224 pixels.


\textbf{Sketch Synthesis Methods.} 
We evaluate 8 sketch synthesis methods: 
\textbf{1) Clipasso}~\cite{vinker2022clipasso} synthesizes sketches using a set of vector strokes. 
We generate sketches with 64 strokes and convert vectors to pixels for evaluation.
\textbf{2) Contour, 3) Anime, and 4) OpenSketch} represent three sketch styles in LineDrawings~\cite{chan2022learning}.
These styles are respectively trained on The Contour Drawings dataset~\cite{li2019photo}, The Anime Colorization dataset~\cite{animedataset2020}, and OpenSketch~\cite{gryaditskaya2019opensketch}.
\textbf{5) PhotoSketch}~\cite{li2019photo} employs a conditional GAN method to predict salient contours in reference photos. 
\textbf{6) UPDG1, 7) UPDG2, and 8) UPDG3} represent three sketch styles in UPDG~\cite{yi2020unpaired}. The styles are respectively trained on images from different artists and illustration websites.

\textbf{Correlation Strength Measurements.}
We use two rank correlation coefficients to evaluate the alignment of various metrics with human perception: Spearman's rank correlation coefficient \(\rho\)~\cite{spearman1961proof} and Kendall's rank correlation coefficient \(\tau\)~\cite{kendall1938new}. 

\textbf{Implementation Details.}
For category-level recognizablity, we utilize CLIP ViT-B/32~\cite{radford2021learning}. 
For structural-level recognizability, we use the pre-trained RTMPose~\cite{jiang2023rtmpose} model from MMPose toolkits for keypoint detection. We implement all the models based on the PyTorch framework and conduct model inference using one NVIDIA RTX-2080Ti GPU.
Additional details are provided in the supplementary materials.

\subsection{Collecting Human Assessment}
To investigate human assessments of sketches on structure-level recognizability and simplicity, we conduct user experiments. 
Considering that sketching is a form of artistic expression, we recruit participants from an art community platform who have received prior artistic training to evaluate the sketches ($\$$20.00/hour). 

The study involved 198 participants, 150 of whom have experience in drawing sketches. 
We select 111 sets of human body sketches from SketchRef, each set containing sketches produced by 5 methods (CLIPasso~\cite{vinker2022clipasso}, Photosketch~\cite{li2019photo}, UPDG~\cite{yi2020unpaired}, Anime~\cite{chan2022learning} and OpenSketch~\cite{chan2022learning}), resulting in a total of 555 sketches.
Participants are randomly provided with 8 sets of body sketches, with 4 sets each dedicated to evaluations of recognizability and simplicity.
For structure-level recognizability evaluation, participants rate the sketches on a 5-point scale based on their ability to identify human postures.
Given that simplicity is a relative concept, we use a ranking system and calculate rank scores based on weighted averages:
\begin{equation}
\label{eq: rank score}
\text{Average Rank Score} = \frac{\sum (\text{frequencies} \times \text{weights})}{\text{number of responses}},
\end{equation}
where weights are assigned based on the ranking position of the options.
In our study, in a ranking of five options, weights are assigned from 5 to 1.
In total, we collected 198 \(\times\) 40 = 7,920 responses, ensuring that each sketch is evaluated by at least three independent participants.



\subsection{Quantitative Results}
\begin{table}[htbp]
\centering
\caption{Comparison between the \(\text{R}_{s}\) obtained by pose estimation models and other similarity metrics, testing their alignment with user-assessed structure-level recognizability.}
\label{tab:align:recognizability}
\begin{tabular}{cccc}
\toprule
\textbf{Paradigms}           & \textbf{Methods}                   & \textbf{$\rho$} & \textbf{$\tau$} \\ \midrule
\multirow{2}{*}{Traditional} & SSIM~\cite{wang2004image}          & 0.5418          & 0.4420          \\
                             & MS-SSIM~\cite{wang2004image}       & 0.3128          & 0.2282          \\ \midrule
\multirow{2}{*}{Perceptual}  & IS~\cite{salimans2016improved}     & 0.1679          & 0.1666          \\
                             & LPIPS~\cite{zhang2018unreasonable} & 0.3262          & 0.2458          \\ \midrule
Keypoints                    & $\text{R}_{s}$(OURS) & \textbf{0.6619} & \textbf{0.5533} \\ \bottomrule
\end{tabular}
\end{table}

\textbf{Consistency of Structural Recognizability with Human Assessment.} We evaluate the effectiveness of \(\text{R}_{\text{s}}\) by comparing its rank correlation with user-assessed structure-level recognizability against four common metrics. As shown in Table~\ref{tab:align:recognizability}, traditional metrics, relying on pixel-level differences, and perceptual metrics, using overly generalized features, fail to capture structural characteristics and show minimal correlation with human assessments. 
In contrast, \(\text{R}_{\text{s}}\) demonstrates strong alignment with human perception, highlighting its superiority.
\begin{table*}[htbp]
\centering
\caption{Benchmark Results. We use \(\text{mRS}@\alpha\) to evaluate these synthesis methods. \(\alpha\) represents the simplification threshold, and "-" represents a value of 0.}
\label{tab:benchmark result}
\begin{tabular}{ccccccccccccc}
\toprule
\multirow{3}{*}{\textbf{Method}} & \multicolumn{6}{c}{\textbf{Structural Consistency Estimation}}                                                                                           & \multicolumn{4}{c}{\textbf{Category Prediction}}                                                   & \multicolumn{2}{c}{\multirow{2}{*}{\textbf{Average}}} \\ \cline{2-11}
                                 & \multicolumn{2}{c}{\textbf{Human}} & \multicolumn{2}{c}{\textbf{Face}} & \multicolumn{2}{c}{\textbf{Animal}} & \multicolumn{2}{c}{\textbf{Animal}} & \multicolumn{2}{c}{\textbf{Things}} & \multicolumn{2}{c}{}                                  \\ \cline{2-13} 
                                 & \textbf{@0}         & \textbf{@1.5}      & \textbf{@0}         & \textbf{@1.5}     & \textbf{@0}          & \textbf{@1.5}      & \textbf{@0}          & \textbf{@1.5}      & \textbf{@0}          & \textbf{@1.5}      & \textbf{@0}               & \textbf{@1.5}            \\ \hline
        Anime~\cite{chan2022learning} & \textbf{94.36} & - & \textbf{72.86} & 4.75 & \textbf{79.07} & - & 72.10 & - & 68.44 & 34.58 & \textbf{77.37} & - \\ 
        UPDG3\cite{yi2020unpaired} & 89.59 & 0.08 & 66.44 & 53.64 & 70.80 & 1.85 & 68.39 & 2.57 & 64.24 & 50.09 & 71.87 & 21.64 \\ \
        UPDG2\cite{yi2020unpaired} & 89.19 & 5.27 & 63.99 & \textbf{63.34} & 56.19 & 23.46 & 69.44 & 33.79 & 64.58 & 61.32 & 68.68 & 37.43 \\ 
        UPDG1\cite{yi2020unpaired} & 89.26 & 36.03 & 58.36 & 58.36 & 57.85 & \textbf{54.20} & 68.16 & 64.54 & 63.04 & 61.54 & 67.32 & 54.93 \\ 
        OpenSketch~\cite{chan2022learning} & 84.75 & - & 56.71 & 0.74 & 54.71 & - & 68.86 & - & 64.82 & 23.11 & 65.97 & - \\ 
        CLIPasso\cite{vinker2022clipasso} & 88.38 & \textbf{88.38} & 43.31 & 43.31 & 47.03 & 47.03 & \textbf{72.60} & \textbf{72.60} & \textbf{69.06} & \textbf{69.06} & 64.08 & \textbf{64.08} \\ 
        Contour~\cite{chan2022learning} & 78.56 & 0.48 & 49.84 & 43.30 & 41.71 & 4.47 & 71.39 & 7.99 & 66.48 & 61.09 & 61.60 & 23.47 \\ 
        PhotoSketch\cite{li2019photo} & 74.82 & 74.82 & 32.16 & 32.16 & 26.58 & 26.58 & 71.38 & 71.38 & 64.37 & 64.37 & 53.86 & 53.86 \\ 
\bottomrule
\vspace{-5mm}
\end{tabular}
\end{table*}

\begin{figure}[htbp]
  \centering
  \subfigure[]{
    \label{fig:erase vis} 
    \includegraphics[height=0.23\textwidth]{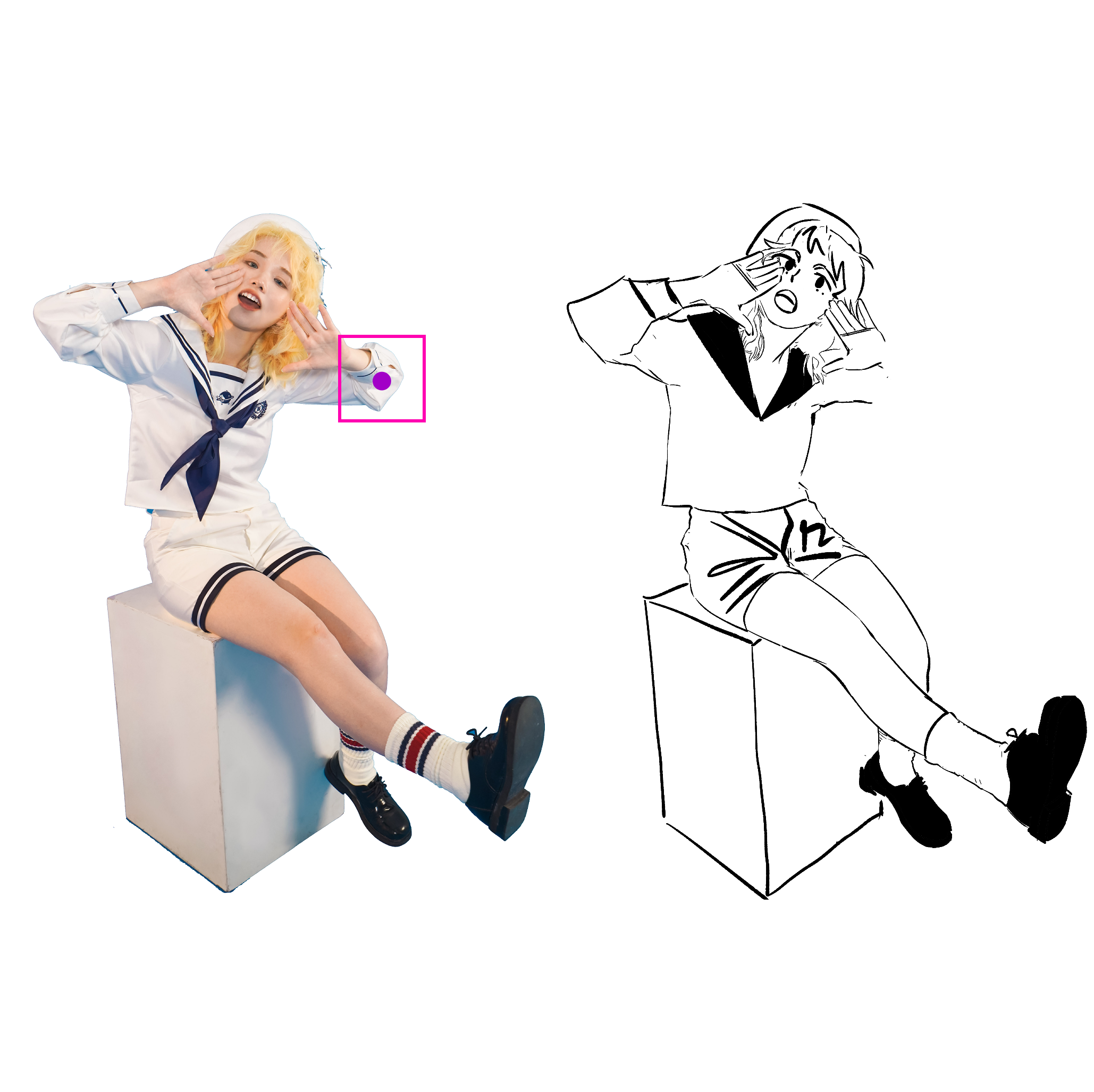}}
  \hspace{0in} 
  \subfigure[]{
    \label{fig:erase line chart} 
    \includegraphics[height=0.23\textwidth]{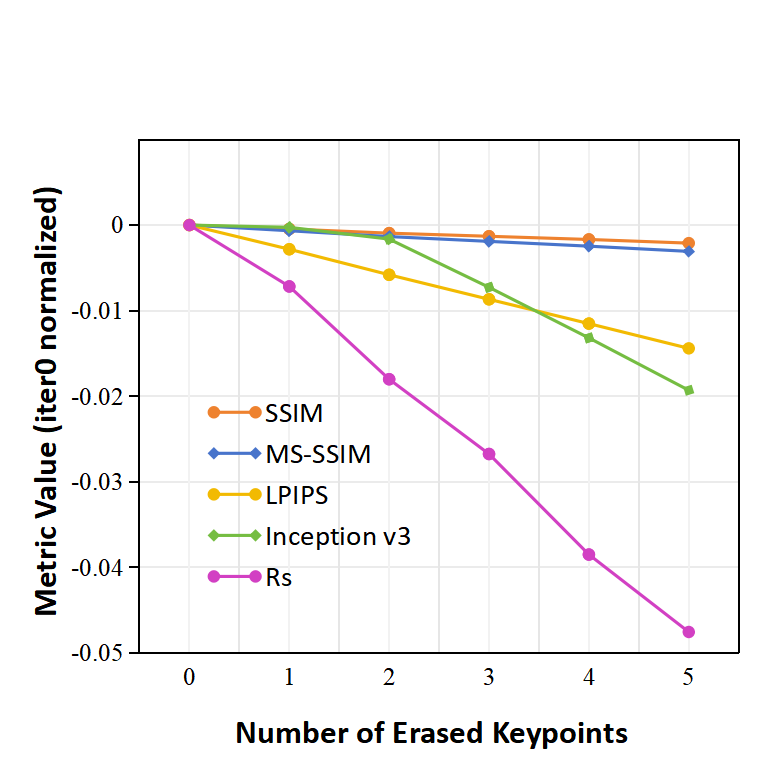}}
  \caption{(a) Example of essential regions: We argue that lines near key points influence the expression of structure, and their erasure can impair the recognition of limb positions.
(b) In the sketches synthesized by Clipasso, we erase a certain number of essential regions and calculate the scores of various similarity metrics on the erased sketches. These scores are normalized by subtracting the scores of the sketches without erasure.}
  \label{fig:erase} 
\end{figure}

\begin{table}[htbp]
\centering
\caption{The alignment between the SR metric calculated by different complexity assessment methods, and user-assessed recognizability.}
\label{tab:align:simplicity}
\begin{tabular}{lcc}
\toprule
\textbf{Complexity Methods in SR}                 & \textbf{$\rho$} & \textbf{$\tau$} \\ \midrule
ICNet~\cite{feng2022ic9600}                & 0.4705          & 0.3734          \\
1d Entropy~\cite{corchs2016predicting}                       & 0.5691          & 0.4062          \\
Fast Corner Detection~\cite{rosten2006machine}            & 0.6315          & 0.4665          \\
2d Entropy~\cite{corchs2016predicting}                       & 0.6512          & 0.4792          \\
Harris Corner Detection~\cite{harris1988combined}          & 0.7441          & 0.6170          \\ \midrule
Compression Ratio(OURS)~\cite{machado2015computerized} & \textbf{0.7618} & \textbf{0.6422} \\ \bottomrule
\vspace{-8mm}
\end{tabular}
\end{table}
\textbf{Sensitivity to Essential Regions.} 
To test the sensitivity of different metrics to the absence of critical structure in sketches, we consider the regions around keypoints as essential regions and erase a certain number of $10\times 10$ pixel areas centered around randomly selected keypoints in the generated sketches (see Fig.~\ref{fig:erase vis}). 
These erased sketches are evaluated using various metrics. 
As shown in Fig.~\ref{fig:erase line chart}, the \(\text{R}_{s}\) metric is particularly sensitive to the erasure of essential regions. 
As the number of erased essential regions increases, \(\text{R}_{s}\) decreases accordingly. 
In contrast, other metrics such as SSIM and MS-SSIM show minimal variation, and the IS score does not decrease when the first two essential regions are erased. 
Although LPIPS does decrease, the change is not substantial. 
This indicates that when the critical structure of a sketch is omitted, \(\text{R}_{s}\) effectively reflects the degradation in sketch quality.

\textbf{Consistency of Simplicity Metrics with Human Assessment.} We also analyze the correlation between the SR metric calculated by different image complexity assessment methods and user-assessed simplicity. 
As shown in Table~\ref{tab:align:simplicity}, the Compression Ratio algorithm~\cite{machado2015computerized} shows a high correlation with human perception ($\rho$=0.7618, $\tau$=0.6422). 
In contrast, although ICNet~\cite{feng2022ic9600} is trained with paintings included in its dataset, the majority of its training data consists of real photographs, which limits its generalization capabilities for sketches. 
Considering that sketches are a relatively simple form of visual expression, we believe that employing the Compression Ratio method in SR for measuring simplicity is sufficient.

\subsection{Benchmark Results and Analysis}

We report the benchmark results for various sketch synthesis methods in Table~\ref{tab:benchmark result}. Based on these results, we have the following insights:

\textbf{A method that excels in category prediction does not necessarily guarantee consistency in key structural features.} 
This is because category conditions are more lenient—for instance, when drawing a cat, the requirement is simply to make it identifiable as a cat. Even if the cat is missing limbs, it is still recognized as a cat. 
In contrast, structural requirements are more stringent, demanding that key parts of the reference image be depicted with essential lines.

\textbf{As the simplification threshold increases, the performance of many synthesis methods tends to degrade.} 
When the simplification threshold is set to 0, Anime~\cite{chan2022learning} and UPDG3~\cite{yi2020unpaired} perform well in both category and structural estimation tasks. 
However, at a threshold of 1.5, their performance may drop significantly, potentially reaching zero. This is because these methods use complex lines to synthesize sketches, attempting to restore unessential details from the reference images while maintaining a low level of simplification.

\textbf{Incorporating semantic loss during training helps improve performance and robustness in category prediction tasks.} 
In the two category tasks, regardless of the simplification threshold, Clipasso~\cite{vinker2022clipasso} consistently outperforms the others. 
This indicates that it maintains category recognizability even during simplification, demonstrating high robustness. 
We believe this is due to its loss function, which incorporates semantic loss.
In contrast, there are currently no methods that consider structural information during training, which results in suboptimal performance in the three structural consistency tasks.

Overall, in the category prediction task, Clipasso~\cite{vinker2022clipasso} achieves the best performance regardless of the simplification threshold. However, in the structural consistency estimation task, no model has yet demonstrated such robustness. Future research should consider incorporating structure-consistent optimization.

\section{Conclusion}
We introduce SketchRef, the first multi-task evaluation benchmark for sketch synthesis.
This benchmark spans multiple domains and conducts evaluations on both category and structure by leveraging the commonalities between sketches and reference photos.
We also propose a general recognizability evaluation principle, mRS, which accounts for the simplicity of sketches, ensuring fairness in the evaluation of sketches with varying levels of simplification. 
Based on SketchRef, we conduct a comprehensive evaluation of 8 representative sketch synthesis methods and provide valuable insights.
We expect this benchmark can guide future sketch synthesis and sketch understanding.

\section*{Acknowledgment}
This work is supported by the projects of  Beijing Science and Technology Program (Z231100007423011) and National Natural Science Foundation of China (No. 62376012), which is also a research achievement of State Key Laboratory of Multimedia Information Processing and Key Laboratory of Science, Technology and Standard in Press Industry (Key Laboratory of Intelligent Press Media Technology).

\bibliographystyle{IEEEbib}
\bibliography{reference}

\end{document}